%% file: main.tex
\pdfoutput=1

\documentclass[11pt]{article}

\usepackage[]{acl}
\usepackage{multirow}
\usepackage{hyperref}
\hypersetup{
    colorlinks=true,
    linkcolor=blue,
    filecolor=magenta,      
    urlcolor=blue,
    pdftitle={Overleaf Example},
    pdfpagemode=FullScreen,
    }

\usepackage{times}
\usepackage{latexsym}
\usepackage{amssymb}
\usepackage{comment}
\usepackage{booktabs}
\usepackage{array}
\usepackage{longtable}
\usepackage[T1]{fontenc}
\usepackage[]{mdframed}

\usepackage{subcaption}

\usepackage[utf8]{inputenc}

\usepackage{microtype}
\usepackage{afterpage}

\usepackage{xspace}
\usepackage{xcolor}

\usepackage{amsmath}
\usepackage{algorithm}
\usepackage{algpseudocode}
\usepackage{subcaption}

\usepackage{cleveref}

\usepackage{graphicx}
\usepackage[]{todonotes}

\definecolor{royalpurple}{RGB}{136,18,255}
\definecolor{royalblue}{RGB}{0,102,204}

\newcommand{\modeli}{\ensuremath{m_{i}}\xspace}
\newcommand{\modelone}{\ensuremath{m_1}\xspace}
\newcommand{\modeltwo}{\ensuremath{m_2}\xspace}

\newcommand{\methodname}{Agreement-Based Ensembling\xspace}

\newcommand{\tower}{\textsc{Tower}\xspace}
\newcommand{\llamafull}{LLaMa 3.x\xspace}
\newcommand{\llama}{LLaMa\xspace}

\newcommand{\nllb}{NLLB\xspace}
\newcommand{\mtom}{M2M\xspace}

\newcommand{\shared}{\checkmark}
\newcommand{\stalled}{\ensuremath{\times}}

\title{Token-level Ensembling of Models with Different Vocabularies}

\author{
    \textbf{Rachel Wicks},\textsuperscript{1}
    \textbf{Kartik Ravisankar},\textsuperscript{2}
    \textbf{Xinchen Yang},\textsuperscript{2} 
    \textbf{Philipp Koehn},\textsuperscript{1} \textnormal{and}
    \textbf{Matt Post}\textsuperscript{1,3} \\
    \textsuperscript{1}Johns Hopkins University \quad 
    \textsuperscript{2}University of Maryland \quad 
    \textsuperscript{3}Microsoft \\
    \texttt{rewicks@jhu.edu, kravisan@umd.edu, xcyang@umd.edu}
}

\begin{document}
\maketitle
\begin{abstract}
Model ensembling is a technique to combine the predicted distributions of two or more models, often leading to improved robustness and performance.
For ensembling in text generation, the next token's probability distribution is derived from a weighted sum of the distributions of each individual model.
This requires the underlying models to share the same subword vocabulary, limiting the applicability of ensembling, since many open-sourced models have distinct vocabularies.
In research settings, experimentation or upgrades to vocabularies may introduce multiple vocabulary sizes.
This paper proposes an inference-time only algorithm that allows for ensembling models with different vocabularies, without the need to learn additional parameters or alter the underlying models.
Instead, the algorithm ensures that tokens generated by the ensembled models \textit{agree} in their surface form.
We apply this technique to combinations of traditional encoder-decoder models and decoder-only LLMs and evaluate on machine translation.
In addition to expanding to model pairs that were previously incapable of token-level ensembling, our algorithm frequently improves translation performance over either model individually.
\end{abstract}

\section{Introduction}
\label{sec:introduction}

\begin{figure}[t]
\centering
\includegraphics[width=0.95\columnwidth]{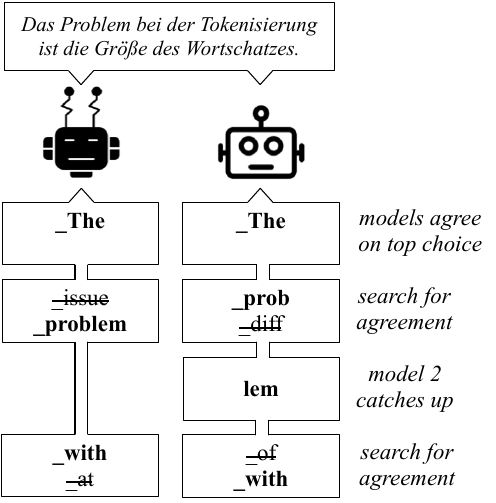}
\caption{\methodname (ABE) enables ensembling among models with different vocabularies.
Token generation for each beam item is constrained to tokens with agreeing detokenized forms.}
\label{fig:intuitive-visual} 
\vspace{-2mm}
\end{figure}

Text generation takes place as a sequence of token predictions.
At each time step, the model, conditioned on some input, produces a probability distribution over the vocabulary.
From this distribution, the next token is selected to extend the hypothesis---the text generated thus far.

Individual models may be sensitive to noise or lack coverage in certain domains.
Model ensembling is a method to combine outputs from multiple models, which often provides for more robust outputs and increases in performance.
The traditional model ensembling approach assumes a shared vocabulary and computes a new distribution as a weighted sum of its component vocabularies:
\begin{equation}
\label{eq:interpolation}
p(x_t) =\sum_i \lambda_i p_{\modeli}(x_t \mid x_{1..t-1})
    \vspace{-2mm}
\end{equation}
where all interpolation weights, $\lambda_i$, are nonnegative and sum to 1.
The new ensembled distribution functions as if it originated from a single model, and the next-token prediction proceeds as usual.

In practice, most models do \textit{not} share vocabularies.
When the vocabularies differ, the resulting probability distributions are no longer combinable.
Then, it is no longer straightforward to ensemble these outputs.
To address this, we introduce \methodname (ABE), an inference-time ensembling algorithm that requires no new parameters or model adaptation, but instead works by coordinating token selection across models under the notion of \emph{agreement} (\S~\ref{section:agreement}).
At each decoding timestep, each model produces its distribution over the next token; our method efficiently searches over their cross-product for tokens that are contextually compatible with the currently generated surface string (\S~\ref{section:enumerating}).
When the tokens are different (but agreeing), the longer token constrains future search (\S~\ref{section:stalling}).
This is caricatured in Figure~\ref{fig:intuitive-visual}.
Our approach easily extends to other inference algorithms such as beam search (\S~\ref{sec:beam}).

Our contributions are as follows.
We
\begin{itemize}
    \setlength\itemsep{0em}
    \item introduce an inference-time algorithm for ensembling models with different vocabularies,\footnote{Our sole requirement is that all models are open-vocabulary and can therefore generate any string.}
    \item demonstrate the ability to ensemble across varying architectures (encoder-decoder, LLMs, or both), and
    \item show improved results in machine translation across a range of models.
\end{itemize}

\noindent Our code is implemented in Python using the Huggingface transformers library \cite{wolf-etal-2019-huggingface} and is open-source.\footnote{\url{https://github.com/mjpost/abe} (Apache-2.0)}

\section{Related Work}

Ensembling is a generally reliable technique for increasing the quality of model outputs that goes back at least as far as \citet{hansen-salamon-1990-neural}.
Although it is more expensive, and therefore often prohibitive in production inference settings, it is useful for example in competitions or for production training scenarios, such as for distillation.
In such settings, the user typically has complete control over model training; ensembled models can be taken from different checkpoints \cite{sennrich-etal-2016-edinburgh} or from completely different training runs initialized from different random checkpoints, and therefore all have the same vocabularies.
\citet{hoang-etal-2024-fly} move a step beyond this by ensembling models with divergent architectures (an MT system and an LLM) and across contexts longer than are supported by all models, but the models still share the same vocabulary.

The situation becomes more difficult when the vocabularies are not shared.
One way to address this is to work at the sequence level instead of the token level.
One such approach is that of \citet{jiang-etal-2023-llm}, who propose LLM-Blender.
It comprises a ranking function that computes pairwise comparisons of complete model outputs and then selects from among them; this approach completely avoids the need to do any kind of token-level ensembling.
\citet{farinhas-etal-2023-empirical} generate multiple translation hypotheses and then explore selecting from among them using voting, minimum Bayes risk, and LLM-based selection.

Sequence-level ensembling has limitations, and the reality of disjoint vocabularies has motivated prior work in token-level ensembling even across different vocabularies.
Existing work, however, requires extra model training.
\citet{xu-etal-2024-bridging} learned mappings across vocabularies that map token representations into a joint space, and employ a variety of filtering methods for efficiency.
\citet{shen-etal-2024-learning} present a ``collaborative decoding'' framework between a lead and assistant model where a classifier dynamically selects which of them will produce the next token at each step of generation; their approach also appears to require a shared vocabulary.

Our work is distinct in that it requires no further training or parameters.
Our approach manages token-level ensembling across different vocabularies at inference time by ensuring that all models in the ensemble agree on the \emph{string} being generated, and interleaves model steps for models that fall behind.

\input{tex-figures/global-hypothesis}

\section{\methodname}
\label{sec:method}

Autoregressive models produce distributions over their vocabularies at each decoding time step.
This process generally continues until the end-of-sequence token is produced or a maximum length is reached.
Greedy decoding, beam search, and sampling are all search algorithms that change how the next token is selected. %

The traditional model ensembling approach (also called here \emph{interpolation-based ensembling}) fits nicely within any of these frameworks, but requires the models to share the same vocabulary.
This approach simply alters the probability distribution to be a weighted sum of the distributions from each model.
Any search algorithm proceeds as before, selecting a token from this new distribution.

When the vocabularies differ, the distributions do not match and we cannot so nicely factor the probability computation from the algorithm.
In \methodname, each model produces its distribution over its own target vocabulary as usual, but algorithmic changes are required to coordinate on the selection of the next token to ensure they agree on the detokenized surface string.
We note that ABE operates by altering the search for the next token and makes no alterations to the underlying model input allowing models to have unique inputs (e.g., prompts) and architectures.

In this section, we will describe these changes.
At a high level, this requires maintaining a shared global agreement state (\S~\ref{section:agreement}), efficiently searching the cross-product of the models' vocabularies (\S~\ref{section:enumerating}), and handling the varying token lengths of the models' differing vocabularies (\S~\ref{section:stalling}).
For ease of presentation, we will describe the algorithm using two models in a greedy decoding setting; this allows us to focus on these new ideas, without the complexity of beam search.
However, the algorithm works with any number of models, so long as they all have open-vocabularies, and the extensions to beam search (which we used for all our experiments) are straightforward.

\subsection{Agreement}
\label{section:agreement}

The fundamental difficulty when ensembling models with different vocabularies is to ensure that they reach consensus on the shared output string, despite the fact that the string will have been generated via different tokenizations.
In \methodname, we maintain a shared string---the global hypothesis---which is updated at each time step by the predicted tokens.
It is important to store and compare against this string in \emph{detokenized} form\footnote{We store byte-strings so byte fall-back tokenization and non-Latin scripts to work.} for precise comparison.
Each model separately maintains its own local hypothesis under its own tokenization, which is a substring of this global hypothesis.
This is visualized in Figure~\ref{fig:global-hypothesis}.

We define the notion of \emph{agreement}.
Consider a set of strings, $S$.
The global hypothesis, $g$, of this set is defined by (1) the shortest terminated string (ends with end-of-sequence token) or (2) the longest unterminated sequence---whichever  is satisfied first.
We define the strings of $S$ to be in \emph{agreement} if and only if all $s_i \in S$ are prefixes of $g$.
Note that agreement does not mean the models have produced the exact same string, only that their strings do not \emph{disagree}.
The algorithm provides a core inductive guarantee that the detokenized string for every model will always agree with the shared global hypothesis.

For the purposes of this paper and all experiments, we require that all models have open vocabularies; i.e., they are able to generate \textit{any} \texttt{utf-8} string.
In modern models, this is typically guaranteed via byte-fallback---using bytes instead of \texttt{<unk>} when a character is not in the vocabulary.
This prevents edge cases while allowing generation in diverse languages.
As long as all models are able to generate the strings of all other models, there will always be at least one set of strings in agreement.

\subsection{Efficient Search}
\label{section:enumerating}

At each decoding timestep, each model takes its forward step from its current state and produces a distribution over its vocabulary.
We need to efficiently search the intersection of their vocabularies for extensions to the current shared hypothesis that are in agreement.
This space has dimensions $V_1 \times V_2$ and is too large to search completely.

We therefore apply a variant of cube pruning \cite{chiang-2007-hierarchical,huang-chiang-2007-forest} with an ``agreement filter'' to search this space efficiently.
The distributions from each model are sorted and arranged into a two-dimensional grid.
This is depicted visually in Figure~\ref{fig:search}.
Each box in the grid denotes the selection of a token from each vocabulary, each of which is associated with a score, computed as the weighted sum of the length-normalized model scores for each local hypothesis.\footnote{In all experiments, models are evenly weighted.} 
As models may be stalled during intermittent time steps, the length of the hypothesis is not guaranteed to be equal to the time step.
This makes length-normalization necessary to prevent bias towards shorter sequences.

To enumerate these items, we maintain a heap, which stores tuple items $(i,j,s)$, where $i$ and $j$ index the candidate vocabulary items, and $s$ records their weighted score.
The neighbors can be enumerated by incrementing exactly one vocabulary index at a time---$(i, j, s_1)$ has two neighbors: $(i+1, j, s_2)$ and $(i, j+1, s_3)$.
Similar to other search algorithms, we do not need to add neighbors that have already been visited (pushed onto the heap).

\input{tex-figures/search-space}

The heap is seeded with the tuple $(1,1,s)$ denoting the top left corner of this grid, representing the most probable token extension from each model.
We now iterate as follows:
    \begin{algorithmic}[1]
        \While{True}
            \State Pop $item$ from $heap$
            \State Compute strings $s_1$ and $s_2$
            \If{agrees($s_1,s_2$)}
            \State{return $item$}
            \EndIf
            \State Add unvisited $neighbors$ of $item$ to $heap$
        \EndWhile
    \end{algorithmic}
Although we need only one valid item for our greedy search example, Figure~\ref{fig:search} depicts the first twelve loop iterations for illustrative purposes.
At each step, the current item is popped from the heap and checked for agreement.
This item is checked to determine whether the set of proposed local hypotheses are in agreement.
Arrows denote ``neighbor'' items (the next vocabulary extension in each dimension), which are used to create updated tuples that are then added to the heap.

This algorithm can be generalized to an arbitrary number of ensembled models by expanding the dimensions of the grid, represented by the expansion of the tuples to include $n$ vocabulary position indices.

\subsection{Stalled steps}
\label{section:stalling}

Models with larger vocabularies are likely to generate longer subwords at each timestep.
This means that one model may be ahead of the rest and need to be \textit{stalled}.
Consider a set of models, $M$.
The set of local hypotheses generated by $M$ is $S$, where $s_i$ was generated by $m_i$.
Recall that the global hypothesis is represented by $g$.
A model, $m_i$, is stalled when $s_i = g$ and at least one other model is not stalled: $\exists (m_j, s_j) \text{ s.t. } s_j \neq g$.
An example of when a model becomes stalled is illustrated in time steps $t=5$ and $t=6$ in Figure~\ref{fig:global-hypothesis}.

Stalled steps aim to restore this imbalance by allowing the unstalled models to generate without the stalled models in order to catch up.
Conceptually, stalling a model is simple.
We prevent the model from being able to generate a token by replacing its vocabulary with an empty transition, $\{\epsilon\}$.
In practice this looks like:
\begin{algorithmic}[1]
\State $\text{scores} \gets []$
\For{$m \in models$}
    \State $seq \gets m.\text{sequence\_score()}$
    \State $s \gets m.\text{step}() + seq$
    \State $s \gets \text{torch.sort}(s)$
    \If{$model.\text{is\_stalled}()$}
        \State $s \gets \{ \text{`values'}: [seq], \text{`idx'}: [-1] \}$
    \EndIf
    \State $\text{scores.append}(s)$
\EndFor
\end{algorithmic}
\input{tex-figures/stalled-search-space}

This uses a dummy vocabulary index (-1) to act as the epsilon transition.
The model's hypothesis score is unaltered as it is not allowed to generate additional tokens.
We illustrate the reduction in search space in Figure~\ref{fig:stalled}.
For each stalled model, the dimensionality of the search space is effectively reduced by one.

\subsection{Beam Search}
\label{sec:beam}

Greedy decoding is a special case of beam search where the beam size is $1$.
It is simple to extend ABE to handle larger beams.
The main conceptual difference is that the search space includes an additional dimension, the beam index.
For a beam size of $k$, the search space is $k~\times~V_1\times~V_2$.
Similar to the extension beyond two models (end of Section~\ref{section:enumerating}), we add an additional index to denote which beam item each vocabulary pair comes from.
Then, instead of terminating after the first valid item, we iterate until we have encountered $k$ of them.
For instance, three models with a beam would have a $4$-dimensional search space of $\{k \times V_1 \times V_2 \times V_3\}$.
The $k$ items become the beam at the next time step.
Note that \textit{neighbors} of a given candidate must come from the same beam item;
beam number $2$ cannot have neighbors in beam number $3$.
This requires that neighbors are \textit{only} increments in exactly one vocabulary dimension---all other dimensions must remain unmodified.

In neural machine translation, beam search typically ensures all beam items are the same length as exactly one token is generated at each time step.
In \methodname, one \textit{or} zero tokens may be generated at each time step.
This is reminiscent of statistical methods which used binning in order to compare hypothesis of equal length.
Rather than binning, we use length-normalization to compare across hypotheses of different lengths.

\section{Experiments}
\label{sec:experiments}

\methodname searches the cross-product of the model vocabularies until an agreed term is found.
If these sorted vocabulary distributions are not similar enough, the algorithm will have to exhaust more of the search space which imposes runtime considerations.

Machine Translation (MT) has a relatively narrow set of acceptable outputs which focuses the output distributions onto similar tokens for any well-trained model.
We therefore choose to evaluate against machine translation as we expect models to agree early and often.
We primarily evaluate on the WMT24 test set \cite{kocmi-etal-2024-findings} \texttt{en-de} but extend to several other out-of-English directions (\texttt{cs}, \texttt{es}, \texttt{uk}) from the same test set.
For evaluation, we consider both COMET \cite{rei-etal-2022-comet} and BLEU \cite{papineni-etal-2002-bleu}.
We computed COMET scores with with pymarian\footnote{Version v1.12.31, wmt22-comet-da model} \cite{gowda-etal-2024-pymarian}, and BLEU scores with sacrebleu\footnote{Version 2.5.1, standard params.} \cite{post-2018-call}.

We examine ensembling within and between different classes of models:
\begin{itemize}
    \setlength\itemsep{0.25em}
  \item\textbf{Custom MT}. We train our own encoder-decoder models on the same pool of data with different vocabulary sizes.
  \item\textbf{Public MT}. Large-scale, multilingual, publicly-available MT models.
  \item\textbf{LLMs}. Decoder-only LLMs with demonstrated capabilities in MT.
\end{itemize}

\subsection{Models}
\label{sec:models}
For preliminary experiments, we start by ensembling models that we trained.
This allows us to have control over the vocabulary while also guaranteeing the models are reasonably similar and will frequently agree during generation.
We then extend to off-the-shelf models, covering both encoder-decoder and decoder-only architectures.

\paragraph{Custom MT} We train transformer base models using Marian \cite{junczys-dowmunt-etal-2018-marian} on approximately 600m lines of filtered English--German data downloaded using \verb|mtdata| \cite{gowda-etal-2021-many} (details in Appendix~\ref{sec:data}).
We perform standard data filtering to include deduplication, language identifiation, length ratios, and margin-scoring.
We train four unigram-based \texttt{sentencepiece} tokenization models \cite{kudo-2018-subword,kudo-richardson-2018-sentencepiece} with sizes of 8k, 16k, 32k, and 64k.
Using these four tokenizers, we train four associated machine translation models.

Each model is a standard transformer base model \cite{vaswani2023attentionneed} with 6/6 (encoder/decoder) layers, embeddings size 1024, and hidden sizes of 8192.
The entire configuration can be found in Table~\ref{tab:marian-config} in the Appendix.
The data is randomly shuffled for infinite streaming via \texttt{sotastream} \cite{post2023sotastreamstreamingapproachmachine}, so we use logical epochs (1b tokens) rather than exact passes over the training set.
We train for 25 logical epochs on one 24GB Titan RTX.
In our experiments, we use various checkpoints of these models.\footnote{Namely epochs \{1, 5, 10, 15, 20, 25\}}

\paragraph{Public MT} In addition to custom models that only support English and German, we also consider two widely used multilingual MT models, \mtom  \cite{DBLP:journals/corr/abs-2010-11125} and \nllb  \cite{nllbteam2022languageleftbehindscaling} in multiple size and distillation variants.
The former covers 100 languages with a 128k multilingual vocabulary, while the latter covers 202 languages with a 256k multilingual vocabulary.
The huggingface repository ids for all off-the-shelf models are listed in Table~\ref{tab:huggingface} in the Appendix.
We note that Public MT models require an additional language tag on both the source and target to signal the language pair to the model.
Again, ABE is agnostic to input changes as it operates on the token distributions.

\paragraph{LLMs} We consider \tower \cite{alves2024toweropenmultilinguallarge} and \llamafull  \cite{grattafiori2024llama3herdmodels}.
\tower is an LLM specifically fine-tuned for the task of translation whereas \llama is general purpose.
\llama models use a vocab of 128k while \tower uses 32k.
\tower was finetuned with the following prompt:
\begin{mdframed}
\small
\textbf{Translate the following text from English into German. %
\textbackslash n
English: \{source sentence\} %
\textbackslash n
German:}
\end{mdframed}
For \llama models, we use both 0-shot prompts and 3-shot prompts\footnote{3-shot experiments were run on one 80GB A100.} derived from the WMT24 baseline evaluation scripts.\footnote{\url{https://github.com/wmt-conference/wmt-collect-translations}}
Exact verbiage of prompts can be found in Table~\ref{tab:prompts} in the Appendix.
LLMs differ in architecture from the previous settings as they lack an encoder.
This further illustrates that ABE is architecture- and input-agnostic.

\input{tex-figures/comet-custom-mt-baseline}

\subsection{Baselines}
\label{sec:baselines}

To compare the results of our ensembling, we have two baseline generation algorithms.
The first is simple translation: using the individual model without ensembling. 
For the MT models, this is only passing the source input (with some language id tags for the multilingual models) to the huggingface \texttt{generate} function.
For \tower and \llama, we use the huggingface \texttt{pipeline} function with the aforementioned prompts (explicitly listed in Table~\ref{tab:prompts}).

We additionally consider linear interpolation as an ensembling baseline.
In this traditional setting, two models' output distributions can only be interpolated when they are over the same event space (i.e., have the same vocabulary).
We therefore only run this baseline over our custom MT models, making use of different checkpoints along the training trajectories of the different models.

For both baselines and \methodname, we use a beam size of 5 for all models.
We generate with a maximum length of 256 tokens.\footnote{
If a model is stalled at this length, there is no agreed hypothesis and we return an empty string.}

\section{Results}
We demonstrate the effectiveness of our ensembling algorithm by comparing the sequences generated by ABE over the performance of the best individual model.
Given two models $m_i$, $m_j$, the translations produced by either model alone are 
$\texttt{T}_{i} \text{ and } \texttt{T}_{j}$, respectively.
The translations produced by ensembling these two models with ABE are denoted as $\texttt{ABE}_{i, j}$.
We define the delta as:
\begin{equation}
    \Delta \mathcal{S} = \mathcal{S}(\texttt{ABE}_{i, j}) - \text{max}(\mathcal{S}(\texttt{T}_{i}), \mathcal{S}(\texttt{T}_{j}))
    \label{eq:metrics}
    \vspace{-0.5mm}
\end{equation}
where $\mathcal{S}$ may refer to BLEU or COMET scores.

\subsection{Custom MT Models}

In Figure \ref{fig:comet-custom-mt}, we display the $\Delta$COMET scores across various combinations of custom MT models.
We provide the $\Delta$BLEU for all custom models in Appendix Figure~\ref{fig:bleu-complete-grid}.
We see consistent positive improvements across many checkpoints. 
First, we consider model combinations which are already compatible with traditional ensembling as a proof of concept.
This is reflected in Figures~\ref{subfig:comet-8kmodels} and \ref{subfig:comet-64kmodels} where the ensembled models have the same vocabulary.
We ensemble the smallest and largest custom MT models with vocabulary sizes of 8k and 64k, respectively, across various checkpoints.
In both cases, we find consistent improvements when using ensembling.

Next, we test the new capability of ABE---ensembling models with differing vocabularies and display the results in Figure~\ref{subfig:comet-8k+64kmodels}.
Not only are we able to successfully ensemble these differing models (a previously impossible task), but we also find positive improvements in COMET scores.

A persistent trend we find is that under-fitted models (e.g., Ep.~1) do not ensemble well.
This is evidenced by negative $\Delta$COMET scores across the first row.
In all other combinations, we see improvement, thus demonstrating the power of ensembling via ABE over using individual models.

We also seek to demonstrate that these ensembling results are at least as good as a naive interpolation-based ensembling baseline.
In order to do this, we compare the relative improvement using interpolation-based ensembling to the improvement gained from ABE.
Note that this restricts the setting in which we can ensemble as the vocabularies \textit{must} match.
In Table~\ref{tab:bleu-interpolation}, we display the relative $\Delta$BLEU improvements and see that ABE is often a bigger improvement in these models.

\input{tex-figures/bleu-interpolation}

\subsection{Public MT Models}\label{subsec:off-the-shelf-MT}

Our custom models are well-suited for ABE, since they were trained on the same data and potentially have related vocabulary distributions even when their vocabularies differ.
We next consider models over which we have less control.
As large multilingual models, \mtom and \nllb are quite different from our custom ones.
In Figure~\ref{fig:off-the-shelf-mt}, we display $\Delta$BLEU.
ABE creates positive improvements though not across all combinations as seen in Custom MT.
\input{tex-figures/off-the-shelf-mt}

We see an improvement when ensembling our largest custom model (64k) with larger multilingual MT models.
We note that the smaller multilingual model (M2M 418M) has a significantly worse BLEU score than the alternatives which may be an unreasonable ensembling combination.
The $\Delta$COMET scores (displayed in Figure~\ref{fig:comet-complete-grid}) with ABE are more negative than their BLEU equivalents.
This suggests that while ABE might be more effective at surfacing specific $n$-grams, it could negatively impact other aspects, such as fluency, which COMET or other neural metrics may penalize.

\subsection{Off-the-Shelf LLMs}
We demonstrate the flexible nature of the algorithm by extending our ensembling results to distinct architectures---encoder-decoder with decoder-only LLMs. 
In Figure~\ref{fig:off-the-shelf-llm}, we display $\Delta$BLEU improvements.
In this section, we only present 3-shot experiments with \llama but all results are available in Figure~\ref{fig:bleu-complete-grid} in the Appendix.

\input{tex-figures/off-the-shelf-llm}

We still see consistent positive gains from ensembling models---particularly when ensembling the bilingual models with the larger multilingual models.
One crucial trend we notice is that poorer performing models, such as the smaller instances of \mtom or \llama, get consistent \textit{negative} results.
This indicates that poorer performing models will only deteriorate the performance of the better model, which is also typical of other ensembling approaches. 
However, we see improvements when ensembling across architectures: +2.7 BLEU when ensembling a small bilingual model with Tower or LLaMa.
We further see improvements when ensembling two LLMs (+1.4 with Tower and LLaMa8b).
As before, we observe more negative results when using COMET (Figure~\ref{fig:comet-complete-grid} in Appendix).

We see here that this new ability to ensemble models with differing vocabularies does not work in all settings.
This new algorithm provides the framework to spur further research which could provide better recommendations for model combinations or lambda weights which we leave to future work.

\subsection{Additional languages}

We additionally study the ensembling of these models with ABE by comparing the performance in other languages (\texttt{cs}, \texttt{es}, \texttt{uk}).
We compare NLLB, Tower, and LLaMa and display the results in Table~\ref{tab:bleu-extra-languages}.
Similar to before, we notice mixed performance across model pairs and target languages.
We suspect this is due to underlying model differences.

\input{tex-figures/bleu-extra-languages}

Tower and LLaMa, which have been a consistently successful ensembling pair, see improvements in all three of these languages.
We further note that according to their respective documentation, neither model supports \texttt{cs} or \texttt{uk},\footnote{There was likely substantial amounts of these languages in the pretraining data.} but we see improvements in both using ABE.

\section{Analysis}
We seek to answer why our ensembling is successful in some scenarios, though not all.
ABE provides a new framework to more thoroughly investigate model ensembling---particularly in combinations that were previously impossible.
There are many avenues to consider, including something as simple as selection of lambda weights.
As a preliminary path of investigation for future work, we provide both a quantitative and qualitative study.

\subsection{Model Preference}
One effect we wish to disentangle is whether this algorithm is an improvement on the search space or on the modeling.
As previously mentioned, interpolation-based ensembling only affects the intermediate token probabilities (a modeling change) and makes no changes to the search procedure.
ABE does a bit of both by severely altering the search and mildly altering the modeling (scoring by the weighted sum of two models instead of one).

To answer this question, we seek to quantify the preferences of each translation under each model.
Given $m_1$, $m_2$ and the associated translations $\texttt{T}_1$, $\texttt{T}_2$.
We can ensemble these models with ABE to generate $\texttt{ABE}_{m_1,m_2}$.
We then determine the ranking of these three translations under each modeling scheme---$m_1$ and $m_2$---by comparing the models' likelihoods of each translation.
In Table~\ref{tab:preference}, we see that models that ensemble well together (top, custom models) also consistently rank the ABE output as the most likely.
They also agree on the most likely output 86\% of the time.
Conversely, we see more mixed preference with M2M and NLLB ($\Delta$BLEU=-0.2) suggesting that ABE cannot overcome underlying modeling disagreements.
This indicates our method is more effectively exploring the search space when models agree.
\input{tex-figures/preference}

\subsection{Constraining Hallucinations}
\label{sec:verbosity}

Standard (same-vocabulary) ensembling can have a normalizing effect on models, for example helping increase their robustness to noise.
Upon examining outputs, we found a recurring trend that ABE also helps prevent models that have begun to hallucinate.
An example is shown in Table~\ref{tab:verbose}.
Here, noisy inputs that are included by design in the WMT24 test sets occasionally trip up individual models, including LLaMa (3B-Instruct-3-SHOT) which responds in German refusing to translate.
Using ABE on all pairs of these models yields the correct output.
This points to the idea that careful selection of ABE model combinations may allow training a small guide model that can be ensembled with a larger model to constrain its bad tendencies.

\input{tex-figures/verbosity-example}

\section{Conclusion}

We have presented an algorithm that enables token-level ensembling of models with distinct vocabularies.
In contrast to prior relevant work, our approach requires no learned mappings of token representations \cite{xu-etal-2024-bridging} or other model fine-tuning.
Instead, we run models in parallel, using a classical approach from parsing and statistical machine translation to efficiently select tokens whose surface representation all models agree on.

We believe the algorithm itself is an interesting contribution to the literature, since it enables (and makes easy) a task that was previously impossible.
Traditional ensembling is a technique that introduces improvements in some, but not all, settings.
It is therefore interesting that our approach also (a) produces gains in a variety of machine translation settings and (b) also often improves over standard ensembling.
Our analysis shows how this variant of ensembling seems to help address search errors in the underlying models, since those models often prefer (as measured by likelihood) the ensembled results to their own selections.

Machine translation was a natural task for this approach.
For one, ensembling is often used to produce higher-quality distilled results.
Second, the translation task helps constraint the generative output to a subset of tokens that meaningful capture the source semantics.
Our agreement-based approach might falter in less constrained tasks.
The implementation is conceptually simple and factored and allows for easy experimentation with different methods for agreement-based search.
We therefore view this as a fruitful topic for future research.

\section*{Limitations and Ethics}
We note a few limitations with our work.
The first is our focus on one task, machine translation.
Machine translation is heavily conditioned on the input, and the accepted translation set is relatively small compared to other tasks.
Though this approach works on Large Language Models, it may not easily extend to other more diverse tasks such as summarization.

We also acknowledge that machine translation is still a generation task, and is prone to the typical generation pitfalls of hallucinations, or erroneous translations---particularly when using LLMs.
Overly relying on error-prone automated translation without a human review can have unintended consequences when used as a means of distributing information.

The authors also acknowledge the assistance of LLMs in the work in this paper---in particular using AI agents like CoPilot and ChatGPT to write code and edit plots.

\bibliography{custom,anthology}
\bibliographystyle{acl_natbib}

\clearpage
\appendix

\section{Appendix}
\label{sec:data}
\input{appendix-tables/en-de-data}
\clearpage

\clearpage

\label{sec:additional-results}

\clearpage
\input{appendix-tables/bleu-mt-custom}

\clearpage
\input{appendix-tables/bleu-complete-grid}

\clearpage
\input{appendix-tables/comet-complete-grid}

\clearpage
\input{appendix-tables/marian-model}

\clearpage
\input{appendix-tables/huggingface-models}

\clearpage
\input{appendix-tables/prompts}

\clearpage

\section{Sampling}
\label{sec:sampling-appendix}
One common use case with autoregressive models is sampling.
As with other search procedures, standard ensembling works transparently with sampling.
As a procedure, sampling is easy to implement with ABE.
Instead of searching over the grid, we sample from each model consecutively (skipping over stalled models).
The vocabulary which we sample from is renormalized to only allow for \textit{agreeing} tokens.

We experimented with adding sampling to \methodname but found that it did not work well.
We hypothesize the instability of sampling with this method stems in some part from the underlying idea that most tokenizers denote whitespace as \textit{leading} (designating word beginnings) and not as \textit{trailing} (designating word endings).
This idea has been shown to have interesting effects on probability distributions \cite{oh2024leadingwhitespaceslanguagemodels}.

As an illustrative example, consider the following German indefinite articles: ``ein'' and ``eine.''
The key difference being that ``eine'' is feminine.
Both of these words are short and fundamental to the German vocabulary, so it is almost guaranteed that both words in their full form are in the model vocabulary.
We further suspect that models with both of these words in their vocabulary have \textit{never} seen ``eine'' tokenized as ``\_ein'' + ``e'' in their training data.

Now consider our previously stated sampling procedure.
Assume from $m_1$, we sample ``\_Eine.''
When conditioned on this decision, we are likely to see \textit{both} ``\_Ein`` and ``\_Eine'' holding most of the probability mass of $m_2$.
Let's assume we sample ``\_Ein'' from $m_2$.
Since the local hypothesis of $m_1$ (``\_Eine'') and the local hypothesis of $m_2$ (``\_Ein'') are in agreement, this is a valid state to be in.
However, when we next sample from $m_2$ to catch up to $m_1$ it is \textit{not} going to have a high probability on ``e'' because it has never seen ``Eine'' tokenized that way during training.

We understand that $m_1$ has implicitly decided to generate the entire word ``Ein'', but it was unable to convey that it was \textit{also} modeling the end of that word due to the tokenization scheme.

Now consider a word-ending tokenization scheme.
Now, $m_1$ samples ``Eine\_'' signifying that it is \textit{done} with this word.
When we constrain the output of $m_2$ on this hypothesis, ``Ein\_'' is \textit{not} going to be sampled because it does not agree.
In order to get into the same predicament, it would need to place high probability on ``Ein'', specifically \textit{not} ending the word which is unlikely if both models wish to generate some version of the word ``a.''

\end{document}

%% file: tex-figures/global-hypothesis.tex
\begin{figure*}[!ht]
\includegraphics[width=\textwidth]{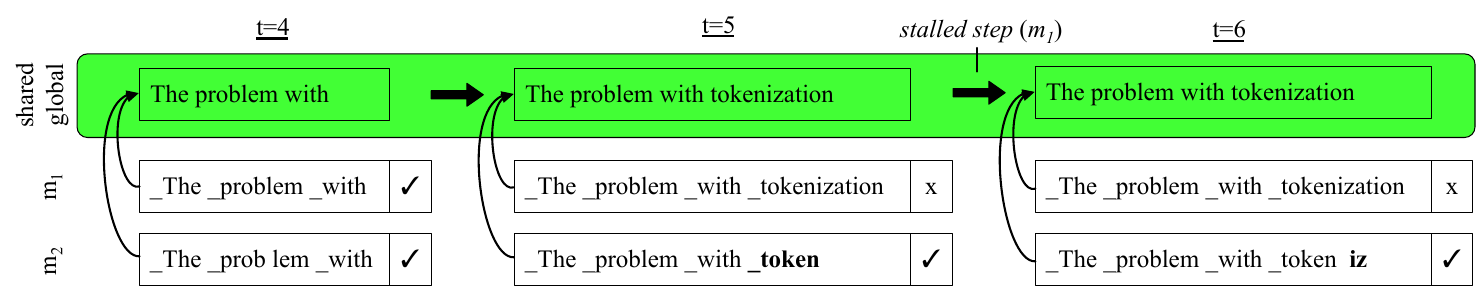}
\caption{
A global state maintains the shared detokenized string, which is determined by the local hypotheses.
Associated with each model is a flag denoting whether the model is stalled (\stalled) or able to generate (\shared).
In stalled steps (\S~\ref{section:stalling}), only the trailing model(s) generate(s) a token, catching up with the shared string.
The stalled model is prevented from generating additional content.
}
\label{fig:global-hypothesis}
\vspace{-2mm}
\end{figure*}

%% file: tex-figures/search-space.tex
\begin{figure}[t]
\includegraphics[width=\columnwidth]{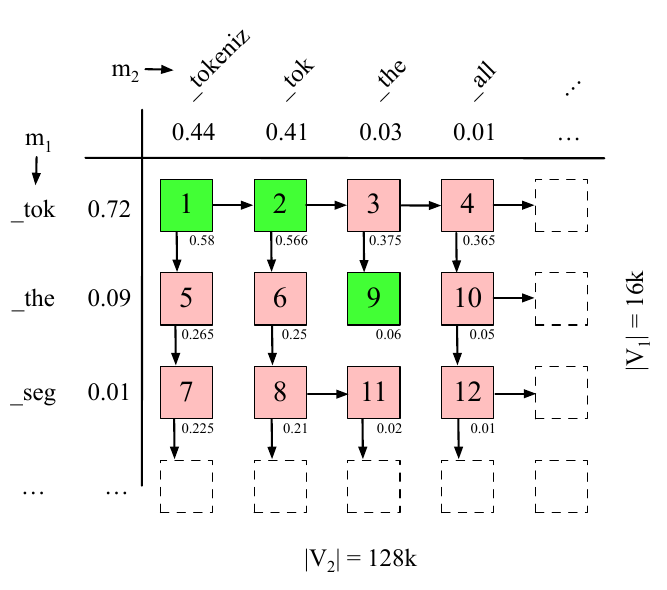}
\caption{
The first 12 candidates in ABE search space for unstalled \modelone, \modeltwo.
Each model's vocabulary is sorted by score.
The top left corner is pushed onto a heap with its weighted score, $0.58$.
We present probabilities here for simplicity.
In practice, each token score is the cumulative log prob of the local hypothesis with this token as the extension.
The loop then pops from the heap, checks for agreement, and adds unvisited neighbors onto the heap.
Numbers denote visitation order.
}
\vspace{-1mm}
\label{fig:search}
\end{figure}

%% file: tex-figures/stalled-search-space.tex
\begin{figure}[h]
\includegraphics[width=\columnwidth]{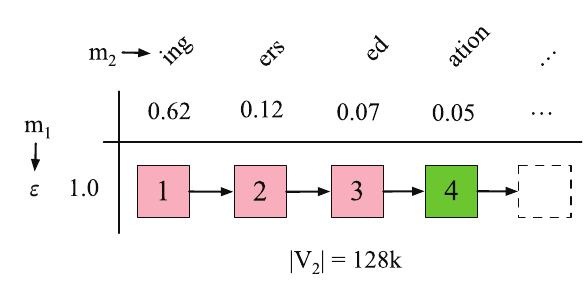}
\caption{
Search space when $m_1$ is stalled.
\modelone has generated \emph{tokenization} while \modeltwo has only generated \emph{\_token iz}.
We present probabilities here for simplicity.
In practice, each token score is the cumulative log-prob of the local hypothesis with this token as the extension.
}
\vspace{-3mm}
\label{fig:stalled}
\end{figure}

%% file: tex-figures/comet-custom-mt-baseline.tex
\begin{figure*}[t!]
    \centering
    \begin{subfigure}{0.33\textwidth}
        \centering
        \includegraphics[width=\linewidth]{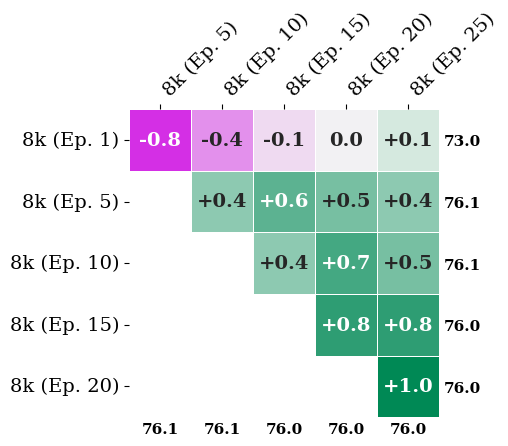}
        
        \caption{Same Vocabulary (Small)}
        \label{subfig:comet-8kmodels}
    \end{subfigure}%
    \hfill
    \begin{subfigure}{0.33\textwidth}
        \centering
        \includegraphics[width=\linewidth]{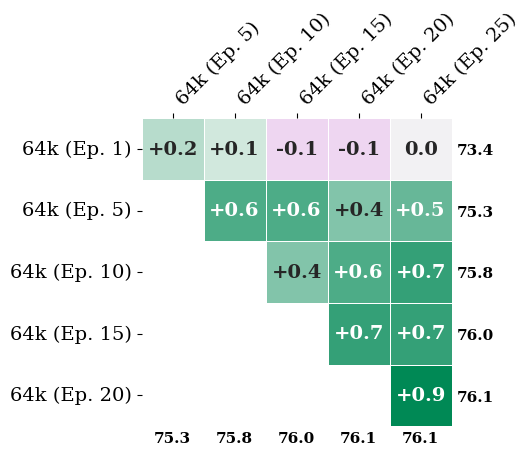}
        \caption{Same Vocabulary (Large)}
        \label{subfig:comet-64kmodels}
    \end{subfigure}
    \hfill
    \begin{subfigure}{0.33\textwidth}
        \centering
        \includegraphics[width=\linewidth]{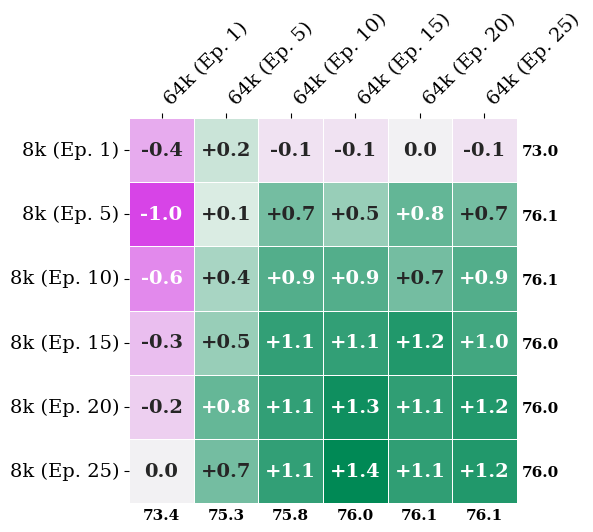}
        \caption{Different Vocabularies}
        \label{subfig:comet-8k+64kmodels}
    \end{subfigure}%
    \caption{$\Delta$COMET results on our custom English–German models using \methodname. $\Delta$COMET is the improvement of ensembling two models via ABE over the best individual model. Individual COMET scores displayed on axes. Labeling indicates vocab size followed by epoch checkpoint. 
    All results on \texttt{en-de} WMT24.}
    \label{fig:comet-custom-mt}
\end{figure*}

%% file: tex-figures/bleu-interpolation.tex
\begin{table}[h]
\centering
\begin{tabular}{@{}l rr@{}}
\toprule
BLEU & $\Delta$Interpolation & $\Delta$ABE \\ \midrule
27.7 & 0.16 & 1.07 \\
\bottomrule
                            
\end{tabular}
\caption{
All scores are averages across all experiments.
We report the average BLEU across models.
For all model pairs
we report the average improvement in BLEU over the score of $m_1$ or $m_2$ individually when using Interpolation or ABE.
}
\label{tab:bleu-interpolation}
\vspace{-2mm}
\end{table}

%% file: tex-figures/off-the-shelf-mt.tex
\begin{figure}[ht]
        \centering
        \includegraphics[width=0.8\linewidth]{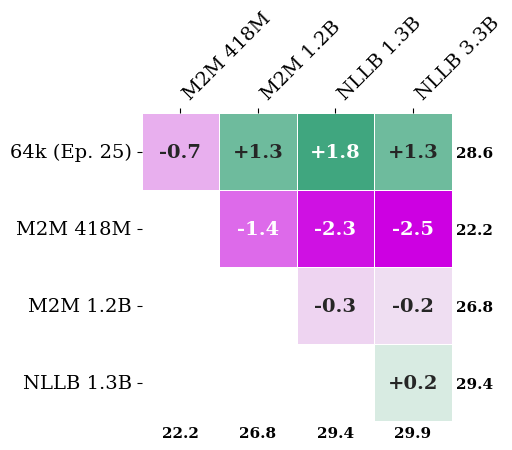}
    \caption{
    $\Delta$BLEU of ensembling different encoder-decoder model pairs using ABE.
    This includes our largest custom model (bilingual) and publicly available multilingual models. 
    Individual BLEU scores displayed on axes. 
    }
    \label{fig:off-the-shelf-mt}
    \vspace{-2mm}
\end{figure}

%% file: tex-figures/off-the-shelf-llm.tex
\begin{figure}[h]
        \centering
        \includegraphics[width=0.8\linewidth]{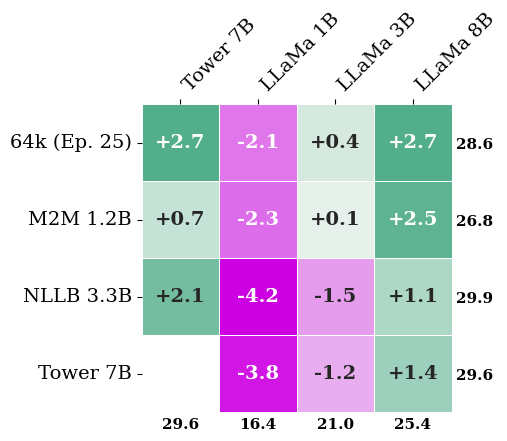}
    \caption{
    $\Delta$BLEU of ensembling various encoder-decoder models with LLMs using ABE. Individual BLEU scores displayed on axes.
    }
    
    \label{fig:off-the-shelf-llm}
\end{figure}

%% file: tex-figures/bleu-extra-languages.tex
\begin{table}[h]
\centering
\small
\begin{tabular}{@{}l l rrr | r@{}}
\toprule
\multicolumn{1}{l}{}
& & $m_1$ & $m_2$ & ABE & $\Delta$ \\ \midrule
\multirow{3}{*}{\scriptsize{\small{NLLB + Tower}}}
& cs &  26.8 &  14.1 &  24.0 &  -2.8  \\
& es & 43.2  & 41.0  &  44.4 & +1.2   \\
& uk & 26.3  &  6.1 &  24.1 & -2.2   \\
\midrule
\multirow{3}{*}{\scriptsize{\small{NLLB + LLaMa}}}
& cs & 26.8  &  19.6 & 25.3  & -1.5   \\
& es & 43.2  & 37.1  & 42.7 &  -0.5  \\
& uk & 26.3  & 20.3  & 26.2  & -0.1   \\
\midrule
\multirow{3}{*}{\scriptsize{\small{Tower + LLaMa}}}
& cs &  14.1 & 19.6  &  21.7  & +2.1 \\
& es & 41.0  & 37.1  & 42.0  &  +1.0 \\
& uk &  6.1 & 20.3  & 22.4  & +2.1 \\
\bottomrule
                            
\end{tabular}
\caption{BLEU scores for different ensembling pairs and their individual models. 
$m_1$ and $m_2$ denote the individual model score while ABE denotes the ensembled score.
$\Delta$ is the difference between ABE and the higher of $m_1$ and $m_2$.
The model versions are M2M 1.2B,
NLLB 3.3B, 
Tower v0.2 7B, 
LLaMa 3.1 8B 3-shot.
}
\label{tab:bleu-extra-languages}
\end{table}

%% file: tex-figures/preference.tex
\begin{table}[h]
\small
\centering
\begin{tabular}{@{}l l rrr | r@{}}
\toprule
\multicolumn{1}{l}{}
& & $\texttt{T}_1$ & $\texttt{T}_2$ & ABE & Same \% \\
\midrule

& & & && \\[-2ex]
\multirow{2}{*}{\scriptsize{\small{8k+64k}}}
& $m_1$ &  102 & 106 & 2207 & \multirow{2}{*}{86.0}\\
& $m_2$ &  198 & 223 & 2028 & \\
\midrule 
& & & && \\[-2ex]

\multirow{2}{*}{\scriptsize{\small{M2M+NLLB}}}
& $m_1$ & 1002 & 1012 & 1092 & \multirow{2}{*}{54.5} \\
& $m_2$ & 840 & 809 & 1096 \\
\bottomrule
                            
\end{tabular}
\caption{Preference. Top: $m_1$ and $m_2$ are our bilingual 8k and 64k models (+$\Delta$ under ABE).
Bottom: $m_1$ and $m_2$ are M2M1.2B and NLLB3.3B (-$\Delta$ with ABE).
$\texttt{T}_i$ shows counts when outputs of $m_i$ were ranked highest (or tied). 
ABE shows counts when the outputs of the ensemble were ranked highest.
``Same \%'' designates when models had the same ranking.
}
\vspace{-2mm}
\label{tab:preference}
\end{table}

%% file: tex-figures/verbosity-example.tex
\begin{table}[ht]
    \centering
    \begin{tabular}{l | p{2.2in}    }
    \toprule
    source
    & lfg \$sqqq
    \\
    \hline
    16k
    & lfg \$sqqq \{m\} \{m\} \{m\} \{m\} \{m\} \dots
    \\
    64k
    & lfg \$qqqq\$qqqqqqqqqqqqqqqqqqq\dots
    \\
    LLaMa
    & Es scheint, dass das ursprüngliche Textstück fehlt oder nicht verfügbar ist. Die gegebene Zeichenkombination "lfg \$sqqq" ist nicht... 
    \\
    \hline 
    ABE 
    & lfg \$sqqq
    \\
    \bottomrule
    \end{tabular}
    \caption{(Truncated) examples of individual models hallucinating or becoming overly verbose on noisy input, but in different ways. The MT models hallucinate with repetition while LLaMa responds in German refusing to translate.
    Any ABE pairing of these models produces the correct output.
    }
    \label{tab:verbose}
    \vspace{-2mm}
\end{table}

%% file: appendix-tables/en-de-data.tex
Below we describe each step of our filtering pipeline:

\begin{enumerate}
    \item Remove items when equal to source and target pair in our validation set.
    \item Remove lines without both source and target.
    \item Remove lines where langid \cite{lui-baldwin-2012-langid} on source is < 0.5 for English and on target is < 0.5 for German.
    \item Remove lines when more than half of the line is punctuation.
    \item Remove lines that have too many characters with frequencies outside of the expected language set \cite{DBLP:journals/corr/abs-2010-11125}.\footnote{\url{https://github.com/facebookresearch/fairseq/blob/main/examples/m2m_100/README.md}}
    \item LASER based Margin-scoring \cite{Artetxe_2019} (done in 2.5M line chunks for computation).
    \item Deduplicate all training data.
\end{enumerate}

\begin{table*}[h]
\centering
    \begin{tabular}{l|r|c}
    \toprule
         Data Name & Filtered Size & Paper (if applicable)  \\
         \hline
         ELRC & 6.5M & \\
         ELRA & 66k & \\
         EU (dcep, eac, ecdc) & 1.8M \\
         Wikimatrix & 5.6M & \citet{schwenk-etal-2021-wikimatrix} \\
         WikiTitles & 2.9M & \\
         TedTalks & 166k & \\
         Bible & 35k & \\
         OPUS Books & 43k & \citet{tiedemann-2012-parallel} \\
         CC-Aligned & 12M & \citet{elkishky2020ccalignedmassivecollectioncrosslingual} \\
         CC-Matrix & 244M & \citet{schwenk-etal-2021-ccmatrix} \\
         DGT & 4M \\
         European Central Book (ECB) & 83k \\
         ELITR & 232k \\
         EMEA & 233k \\
         EU Bookshop & 5.1M \\
         EU Const. & 4k & \\
         Europarl (v3,7,8,10) & 6.3M & \citet{koehn-2005-europarl}\\
         EuroPat (v1-3) & 47M & \citet{heafield-etal-2022-europat} \\
         Global Voices & 174k & \citet{nguyen-daume-iii-2019-global} \\
         JRC & 457k & \citet{steinberger-etal-2006-jrc} \\
         KDE/GNome & 110k & \citet{hatty-etal-2017-creating} \\
         MultiUN & 118k & \citet{chen-eisele-2012-multiun} \\
         MultiCCAligned & 60M \\
         MultiParaCrawl & 70M \\
         News Commentary (v9,14,16) & 937k \\
         OPUS Train & 580k & \citet{tiedemann-2012-parallel} \\
         ParaCrawl (v9) & 242M & \citet{espla-etal-2019-paracrawl} \\
        PHP & 7k & \\
        QED & 400k & \\
        Tanzil & 476k \\
        Tatoeba & 1.8M & \citet{tiedemann2020tatoebatranslationchallenge}\\
        TED (2013) & 403k & \citet{cettolo-etal-2013-report}
        \\
        XLEnt & 1.4M & \citet{el-kishky-etal-2021-xlent} \\
        Tilde & 4.8M & \citet{rozis-skadins-2017-tilde} \\
        StatMT 13 (CommonCrawl) & 1.8M & \\
        \hline
        Deduplicated & 618M & \\
         \bottomrule

    \end{tabular}
    \caption{We aggregate most English–German bitext listed on \texttt{mtdata} (available at \url{https://github.com/thammegowda/mtdata}). The above is the filtered text sizes. }
    \label{tab:data}
\end{table*}

%% file: appendix-tables/bleu-mt-custom.tex
\begin{figure*}[h!]
    \centering
    \begin{subfigure}{0.33\textwidth}
        \centering
        \includegraphics[width=\linewidth]{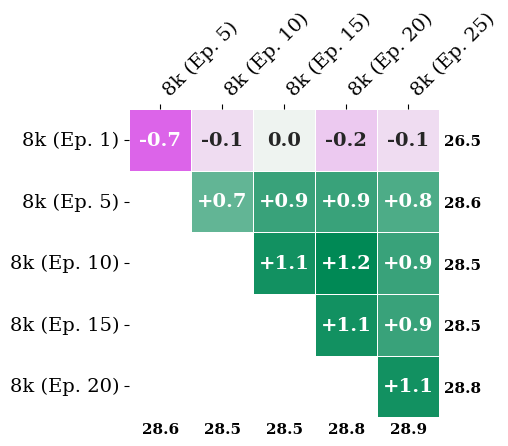}
        
        \caption{Same Vocabulary (Small)}
        \label{subfig:bleu-8kmodels}
    \end{subfigure}%
    \hfill
    \begin{subfigure}{0.33\textwidth}
        \centering
        \includegraphics[width=\linewidth]{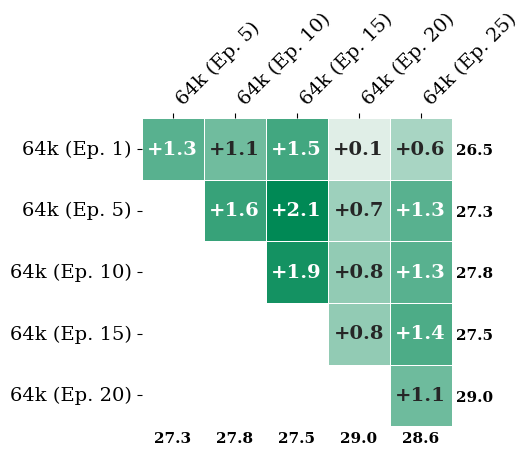}
        \caption{Same Vocabulary (Large)}
        \label{subfig:bleu-64kmodels}
    \end{subfigure}
    \hfill
    \begin{subfigure}{0.33\textwidth}
        \centering
        \includegraphics[width=\linewidth]{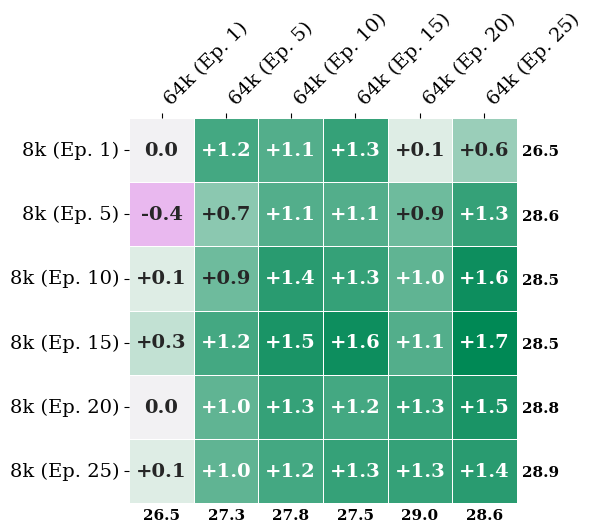}
        \caption{Different Vocabularies}
        \label{subfig:bleu-8k+64kmodels}
    \end{subfigure}%
    \caption{BLEU results on our custom English–German models using \methodname. These charts show the $\Delta$ BLEU improvement of ensembling two models via ABE over the best individual model. Labeling indicates vocab size followed by epoch checkpoint.}
    \label{fig:bleu-custom-mt}
\end{figure*}

%% file: appendix-tables/bleu-complete-grid.tex
\begin{figure*}[t!]
    \centering
    \includegraphics[width=\textwidth]{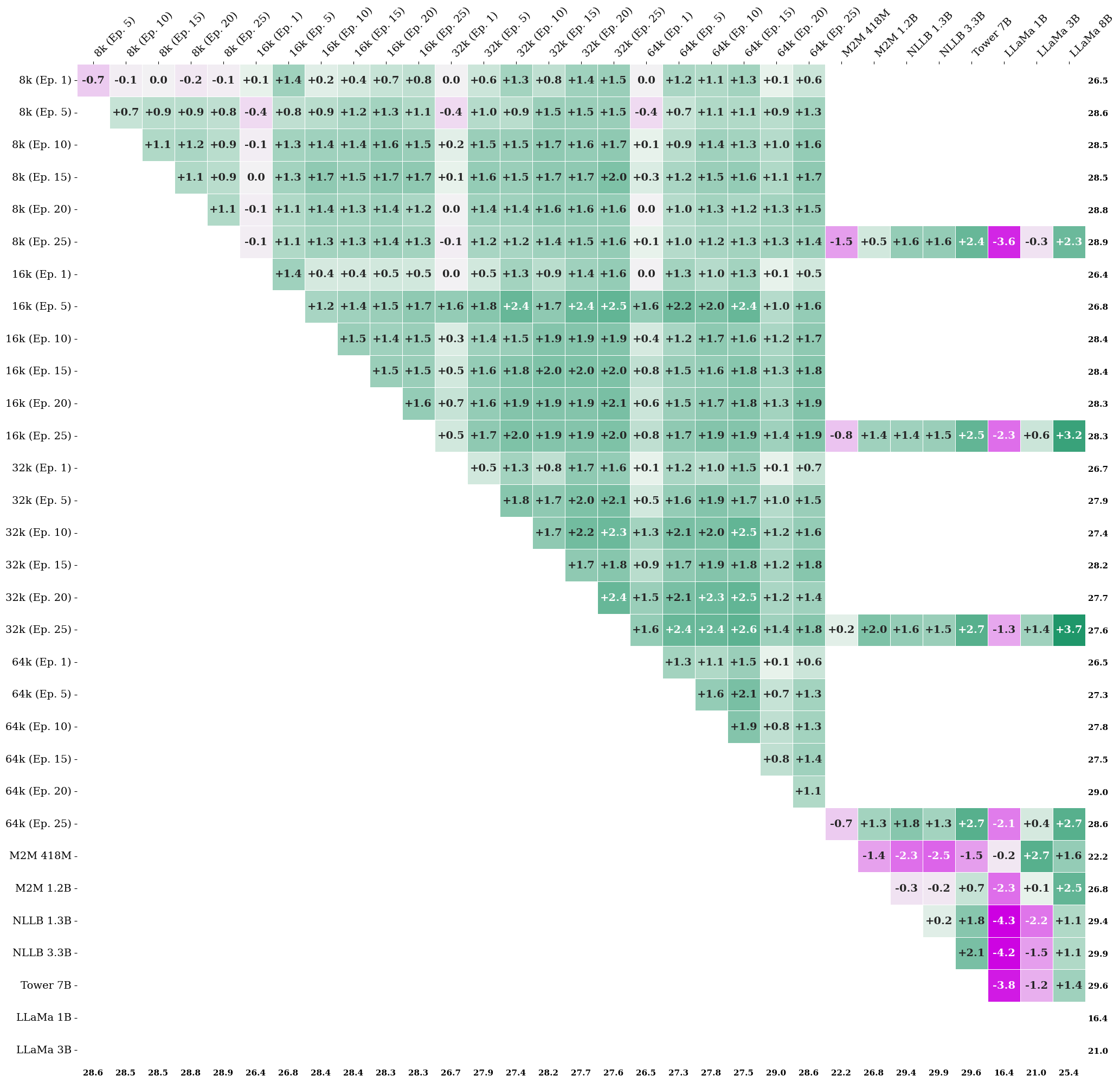}
    \caption{The $\Delta$ BLEU scores for all model pairs.}
    \label{fig:bleu-complete-grid}
\end{figure*}

%% file: appendix-tables/comet-complete-grid.tex
\begin{figure*}[t!]
    \centering
    \includegraphics[width=\textwidth]{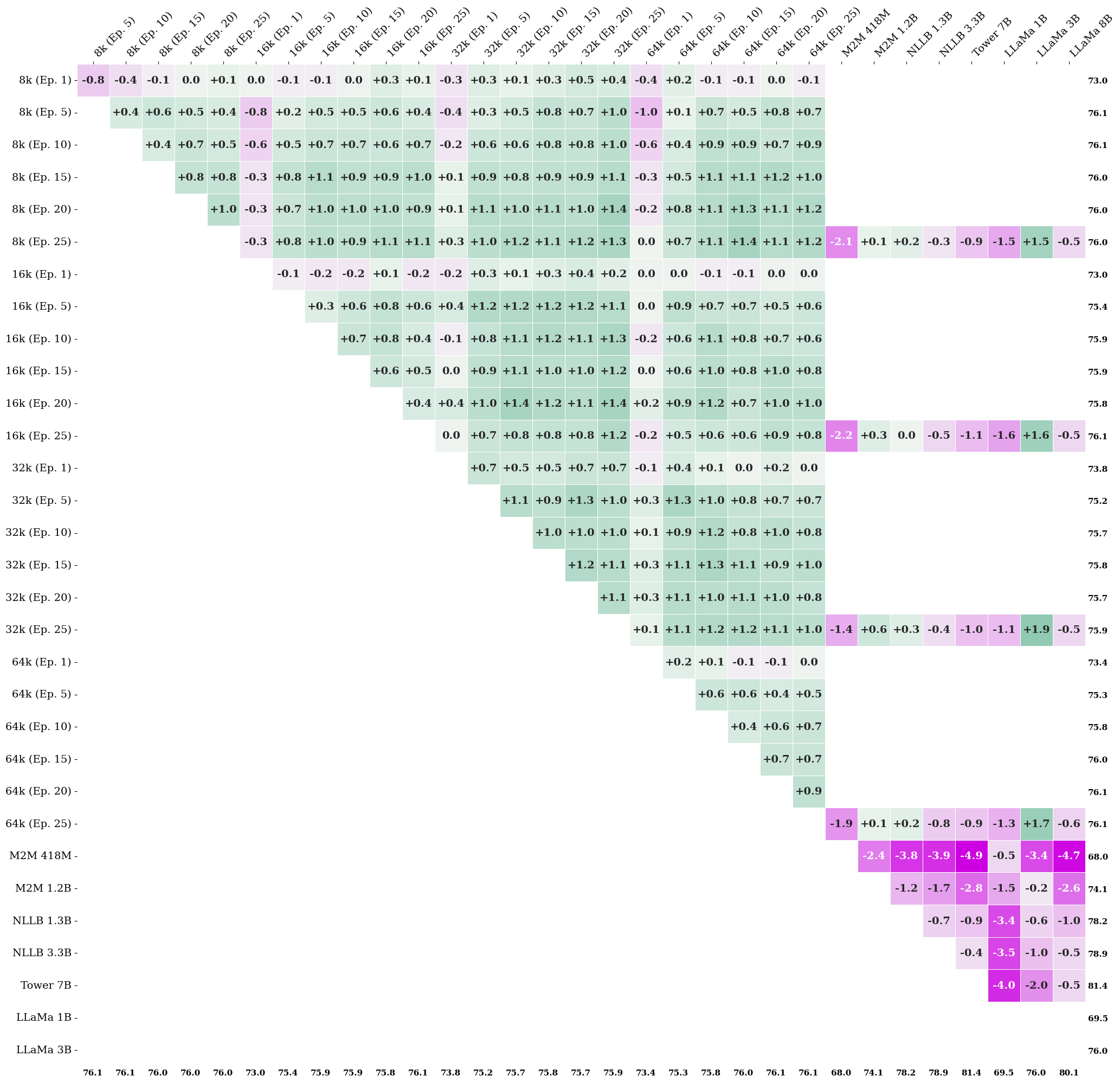}
    \caption{The $\Delta$ COMET scores for all model pairs.}
    \label{fig:comet-complete-grid}
\end{figure*}

%% file: appendix-tables/marian-model.tex
\begin{table*}
\centering
    \begin{tabular}{l | r}
        \toprule
         Hyper-Parameter & Value \\
         \hline
         label smoothing & 0.1 \\
         learning rate &  0.0005 \\
         lr warmup & 4000 \\
         lr decay inv sqrt & 4000 \\
         mini batch warmup & 4000 \\
         
         \hline
        mini batch & 1000 \\
        mini batch words & 500000 \\
        max length & 256 \\
        mini batch fit & true \\
        \hline

        early stopping & 40 \\
        logical epoch &  1Gt \\
        shuffle & batches \\
        fp16 & false \\
        \hline
        tied embeddings & true \\
        tied embeddings all & true \\
        dim emb & 1024 \\ 
        enc depth & 6 \\
        dec depth & 6 \\
        transformer dim ffn & 8192 \\
        transformer decoder dim ffn & 8192 \\
        transformer depth scaling & true \\
        lemma dim emb & 0 \\

        transformer ffn activation & relu \\
        
        transformer-heads & 8 \\
        
        transformer dropout & 0.1 \\
        transformer dropout attention &  0 \\
        transformer dropout ffn & 0.1 \\

        \bottomrule
    \end{tabular}
    \caption{The above enumerate the Marian hyperparameters used for all of our custom models.}
    \label{tab:marian-config}
\end{table*}

%% file: appendix-tables/huggingface-models.tex
\begin{table*}
    \begin{tabular}{|c|>{\raggedright\arraybackslash}p{4.8cm}|c|c|c|>{\raggedright\arraybackslash}p{2.5cm}|}
    \hline
         Model Type & Repo ID/URL &  $m$ Size & $V$ Size & Languages & License \\
         \hline
         \hline
          \multirow{5}{*}{LLM} & \texttt{meta-llama/\allowbreak{}Llama-3.1-8B-\allowbreak{}Instruct} & 8B & 128k & de,es & LLaMa3 \\
          \cline{2-6}
          & \texttt{meta-llama/\allowbreak{}Llama-3.2-1B-\allowbreak{}Instruct} & 1B & 128k & de, es & LLaMa3\\
          \cline{2-6}
          & \texttt{meta-llama/\allowbreak{}Llama-3.2-3B-\allowbreak{}Instruct} & 3B & 128k & de, es & LLaMa3\\
          \cline{2-6}
          & \texttt{Unbabel/\allowbreak{}TowerInstruct-7B-v0.2} & 7B & 32k & de, es & CC-BY-NC-4.0, LLaMa2\\
          \cline{2-6}
          & \texttt{Unbabel/\allowbreak{}TowerInstruct-\allowbreak{}Mistral-7B-v0.2} & 7B & 32k & de, es & CC-BY-NC-4.0, LLaMa2\\
          \hline
          \hline
          \multirow{6}{*}{Public MT} & \texttt{facebook/\allowbreak{}m2m100\_1.2B} & 1.2B & 128k & de, es, cs, uk & MIT \\
          \cline{2-6}
          & \texttt{facebook/\allowbreak{}m2m100\_418M} & 418M & 128k & de, es, cs, uk & MIT \\
          \cline{2-6}
          & \texttt{facebook/\allowbreak{}nllb-200-1.3B} & 1.3B & 256k & de, es, cs, uk & CC-BY-NC \\
          \cline{2-6}
          & \texttt{facebook/\allowbreak{}nllb-200-3.3B} & 3.3B & 256k & de, es, cs, uk & CC-BY-NC \\
          \cline{2-6}
          & \texttt{facebook/\allowbreak{}nllb-200-\allowbreak{}distilled-1.3B} & 1.3B & 256k & de, es, cs, uk & CC-BY-NC\\
          \cline{2-6}
          & \texttt{Facebook/\allowbreak{}nllb-200-\allowbreak{}distilled-600M} & 600M & 256k & de, es, cs, uk & CC-BY-NC\\
        \hline
        \hline
        \multirow{4}{*}{Custom MT\textsuperscript{\textdagger}}& \href{https://huggingface.co/collections/rewicks/baseline-en-de-8k-vocab-670866b90e79a8b46f77a6a0}{rewicks/\allowbreak{}baseline\_en-de\_8k\_ep*} & 286M & 8k & de & Apache 2.0 \\
        \cline{2-6} &
\href{https://huggingface.co/collections/rewicks/baseline-en-de-16k-vocab-6708670f7f0b49b6a7d61251}{rewicks/\allowbreak{}baseline\_en-de\_16k\_ep*} & 294M & 16k & de & Apache 2.0 \\
        \cline{2-6} &    \href{https://huggingface.co/collections/rewicks/baseline-en-de-32k-vocab-6708672f8970d752b70f6e0f}{rewicks/\allowbreak{}baseline\_en-de\_32k\_ep*} & 310M & 32k & de & Apache 2.0 \\
        \cline{2-6} &
        \href{https://huggingface.co/collections/rewicks/baseline-en-de-64k-vocab-670c5bc5ec03e84573b58a8e}{rewicks/\allowbreak{}baseline\_en-de\_64k\_ep*} & 343M & 64k & de & Apache 2.0 \\
        \hline        
    \end{tabular}
    \caption{Huggingface Repo Ids for our publicly available models. LLaMa3 license refers to \url{https://www.llama.com/llama3/license/}. LLaMa2 refers to \url{https://ai.meta.com/llama/license/}. Tower also states the LLaMa license as it uses the LLaMa 2 pretraining weights. Language set only covers those addressed in this paper. \\
    \textsuperscript{\textdagger} Each en-de custom MT model has 40 (epoch 1 to epoch 40) checkpoints, all of which are available in the above-mentioned URL-s.}
    \label{tab:huggingface}
\end{table*}

%% file: appendix-tables/prompts.tex
\begin{table*}
    \begin{tabular}{|c|>{\raggedright\arraybackslash}p{14.5cm}|}
    \hline
    Shot & Prompt \\
    \hline

    0 & [\newline
    \texttt{\{"role": "system", "content": "Cutting Knowledge Date: December 2023\textbackslash nToday Date: 26 Jul 2024"\}} \newline
    \texttt{\{"role": "user", "content": "Translate the following segment into XX. Do not add any additional content. Do not add parentheticals. Only provide the translation. The English segment:"\}} \newline
    ] \\
    
    \hline    

    3 & The example translations are identical to the WMT24 evaluation scripts specific to the target language. The examples can be found at \url{https://github.com/wmt-conference/wmt-collect-translations/tree/main/few_shots}.
    Each example is put in the same format. Language names exchanged when necessary:
    \newline
    
    [\newline
    \texttt{    \{"role": "user", "content": "Translate the following text from English into German. The English Segment: {example source}\}} \newline
    \texttt{    \{"role": "assistant", "content": "\{example translation\}"\}} \newline
    ]
    
    \\
    
    \hline

    \end{tabular}
    \caption{LLaMa prompting messages.}
    \label{tab:prompts}
\end{table*}